\documentclass[runningheads]{llncs}
\usepackage{graphicx}

\usepackage{tikz}
\usepackage{comment}
\usepackage{amsmath,amssymb} 
\usepackage{color}

\usepackage[accsupp]{axessibility}  

\usepackage{cases}
\usepackage{multirow}
\usepackage{wrapfig,lipsum,booktabs}
\usepackage{diagbox}
\usepackage{pifont}

\newcommand{\ie}{\textit{i}.\textit{e}.}
\newcommand{\eg}{\textit{e}.\textit{g}.}

\newcommand{\etc}{\textit{etc}}
\newcommand{\cmark}{\ding{51}}%
\newcommand{\xmark}{\ding{55}}%
\usepackage[pagebackref,breaklinks,colorlinks]{hyperref}

\begin{document}
\pagestyle{headings}
\mainmatter

\title{Collaborating Domain-shared and Target-specific Feature Clustering for Cross-domain 3D Action Recognition} 

\titlerunning{CoDT}
%
\author{ Qinying Liu \and
Zilei Wang\thanks{Corresponding Author}}
\authorrunning{Q. Liu, Z. Wang}
%
\institute{University of Science and Technology of China, Hefei, China \\
\email{lydyc@mail.ustc.edu.cn, zlwang@ustc.edu.cn}}
\maketitle

\begin{abstract}
In this work, we consider the problem of cross-domain 3D action recognition in the open-set setting, which has been rarely explored before. Specifically, there is a source domain and a target domain that contain the skeleton sequences with different styles and categories, and our purpose is to cluster the target data by utilizing the labeled source data and unlabeled target data. For such a challenging task, this paper presents a novel approach dubbed CoDT to collaboratively cluster the domain-shared features and target-specific features. CoDT consists of two parallel branches. One branch aims to learn domain-shared features with supervised learning in the source domain, while the other is to learn target-specific features using contrastive learning in the target domain. To cluster the features, we propose an online clustering algorithm that enables simultaneous promotion of robust pseudo label generation and feature clustering. Furthermore, to leverage the complementarity of domain-shared features and target-specific features, we propose a novel collaborative clustering strategy to enforce pair-wise relationship consistency between the two branches. We conduct extensive experiments on multiple cross-domain 3D action recognition datasets, and the results demonstrate the effectiveness of our method. Code is at \href{https://github.com/canbaoburen/CoDT}{CoDT}.
\end{abstract}

\section{Introduction}
\begin{figure}[t]
\centering
\includegraphics[width=1\linewidth]{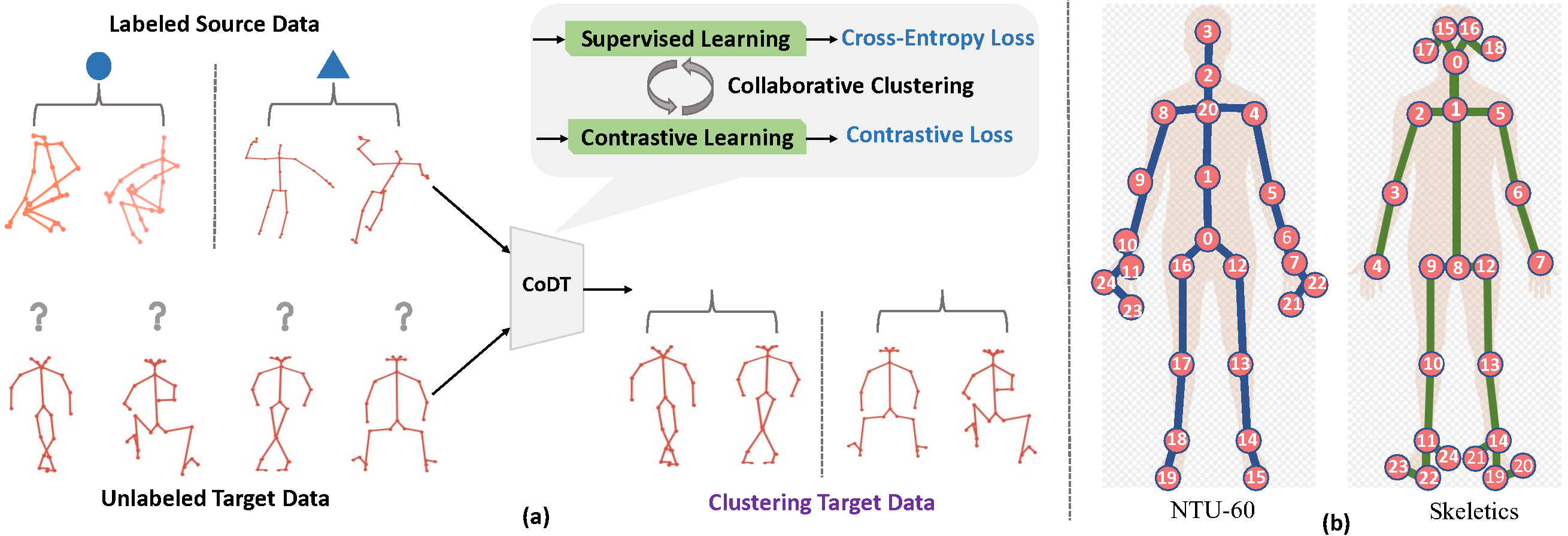}
\centering
\caption{(a) Illustration of CD-SAR task and our proposed CoDT method. CD-SAR aims to group the target samples into semantic clusters by virtue of labeled source data and unlabeled target data. The core idea of CoDT is to collaborate the supervised learning of source data and contrastive learning of target data.
(b) Illustration of different joint definitions in the NTU-60~\cite{shahroudy2016ntu} and Skeletics~\cite{gupta2021quo}. The figure is modified from~\cite{yang2021unik}.}

\label{fig:task}
\end{figure}

Recent advances in 3D depth cameras and pose estimation algorithms have made it possible to estimate 3D skeletons quickly and accurately.
In contrast to RGB images, the skeleton data only contain the coordinates of human keypoints, providing high-abstract and environment-free information. Thus, 3D action recognition (\textit{a.k.a}, skeleton-based action recognition) is attracting more and more attentions~\cite{yan2018spatial,liu2020disentangling,zhang2020semantics}. Nevertheless, existing methods mainly focus on the traditional supervised classification. In this learning paradigm, it is assumed that the labeled training (source) dataset and unlabeled test (target) dataset have the same distribution. In practical scenarios, it is not easy to hold such assumption,
because labeling a dataset with the same distribution as the target data is a laborious task. In reality, it is preferred to utilize a related public annotated skeleton or even image dataset as a source dataset. Unfortunately, there is typically discrepancy (\textit{a.k.a}, domain gap~\cite{ben2010theory}) between the source and target datasets due to various factors, including the devices (\eg, 3D sensors~\cite{zhang2012microsoft,fankhauser2015kinect}, image-based pose estimation algorithms~\cite{cao2017realtime,kocabas2019vibe}), 
the camera setup (\eg, viewpoints), the scenes (\eg, in-the-lab or in-the-wild), \etc. These factors make the skeletons of different datasets distinct in \textit{styles} (\eg, joint types, qualities, views) and action \textit{categories}.
For example, consider a very practical application that needs to recognize the action classes of the in-the-wild unlabeled 3D dataset Skeletics~\cite{gupta2021quo}, we might seek help from the large-scale in-the-lab dataset NTU-60~\cite{shahroudy2016ntu} with high-quality skeletons and annotated labels. However, the NTU-60 and Skeletics are captured by Microsoft Kinect V2 camera~\cite{fankhauser2015kinect} and the pose estimation method VIBE~\cite{kocabas2019vibe} respectively, resulting in different styles of skeletons, \eg, different definitions of human joints (see Fig.~\ref{fig:task}(b)). Furthermore, NTU-60 mainly contains indoor actions, while Skeletics contains more unconstrained actions from internet videos, 
thereby leading to different action categories.
In summary, the domain gap problem is very practical in skeleton-based action recognition but rarely studied in the literature. In this paper, we present the first systematic study to the domain gap problem in skeleton-based action recognition, where the source and target datasets have different \textit{styles} and \textit{categories}. Obviously, this is an open-set problem, and it is expected that the algorithm can automatically cluster the target samples into latent classes.

The labeled source dataset, although collected from a different domain from the target domain, is helpful to develop \textit{domain-shared} representations that can be transferred from the source domain to the target domain. 
By training the models in the standard supervised manner, the representations are expected to be discriminative to different action categories.
However,  the models generally generalize poorly to the target domain due to the domain gap~\cite{yang2021unik,guo2020broader}.
On the other hand, the target domain has many unlabeled samples, which can be used to learn more \textit{target-specific} representations that reveal the inherent characteristics of the target domain. 
This suggests the possibility of applying recently emerging contrastive learning~\cite{he2020momentum,chen2020simple,grill2020bootstrap} on target data. The contrastive learning optimizes the features by instance discrimination, \ie, the features are enforced to be invariant for different transformations of an instance and distinct for different instances. By learning to attract or repel different instances, the features appear to automatically capture some extent of semantic similarity. Yet, they may not have enough discrimination to action categories.

Based on the above discussions, we consider that the domain-shared representations and target-specific representations are conceptually complementary.
This motivates us to integrate the supervised learning on labeled source data and contrastive learning on unlabeled target data.
Previous methods~\cite{phoo2020self,zhong2019invariance,islam2021broad} commonly implement such integration through multi-task learning with a shared feature encoder. However, in our case, the two domains may differ considerably in styles and even joint types, and thus we argue that the domain-shared and target-specific features should be learned via different models in order to sufficiently exploit their respective merits. 
With the purpose of clustering target samples, a natural problem arises: \textit{is it possible to collaborate the two models on feature clustering?} To achieve this goal, we need to address two key issues. The first one is how to cluster the features for a single model, which is expected to be optimized in an end-to-end manner rather than computed by some offline clustering algorithms (\eg, k-means) whose usefulness is proven to be limited~\cite{caron2018deep}.
The second one is how to collaborate both models to jointly optimize feature clustering. It is quite challenging since the learned clusters from two models cannot be matched exactly due to the lack of labels.

In this paper, we propose a novel \textbf{Co}llaborating \textbf{D}omain-shared and \textbf{T}arget-specific features clustering (CoDT) network for the task of cross-domain skeleton-based action recognition (CD-SAR). Fig.~\ref{fig:task} illustrates the overview of the task and our method. Specifically, to address the first issue, we propose an online clustering algorithm to generate robust pseudo labels to guide feature clustering. It is built upon the teacher-student framework~\cite{tarvainen2017mean}, where the clustering is optimized via the pseudo labels generated by the teacher model. Under this framework, a straightforward way to determine pseudo label is to select the cluster with which the teacher is most confident.
However, due to the discrepancy between target sample clustering and source-based supervised learning (or instance-based contrastive learning), there is a risk of obtaining a trivial solution that groups most samples into only a few clusters, as observed in Sec.~\ref{exp:OCM}. 
To make the clusters balanced, we propose to generate uniformly distributed pseudo labels. It is non-trivial to achieve it online as we have to take into consideration the global distribution of pseudo labels. 
Hence, we transform it to an optimal transport problem that can be solved by linear programming~\cite{cuturi2013sinkhorn}.

As for the second issue, we propose a collaborative clustering strategy that exchanges pseudo labels across models for collaborative training (co-training). In the traditional co-training methods~\cite{blum1998combining,qiao2018deep,han2018co}, the categories of two models are pre-defined and consistent, and thus the pseudo labels produced by one model can be directly used to train another model. In our case, however, the semantics of the learned clusters of two models are agnostic and variable during training, making it difficult to determine the correlation between the clusters of two models. To this end, we propose to perform co-training on the pair-wise relationship that represents whether a pair of samples are from the same cluster (positive pair) or distinct clusters (negative pair). Specifically, we first construct the pair-wise binary pseudo labels by comparing the instance-wise pseudo labels of samples. Thereafter, the pair-wise labels are used to train the other model, where a novel contrastive loss is particularly adopted to enforces the model to produce consistent/inconsistent predictions for the positive/negative pairs.

Our contributions are summarized as: 
1) We provide a benchmark for CD-SAR. To solve this task, we propose a novel two-branch framework dubbed CoDT to collaborate domain-shared and target-specific features.
2) We propose an online clustering algorithm that can alternate the robust pseudo label generation and balanced feature clustering.
3) We propose a collaborative clustering algorithm, which enables co-training of two models to enforce their consistency in terms of pair-wise relationship.  
4) We evaluate our method upon different cross-domain tasks, and the effectiveness of our method is well shown.

\section{Related Work} \label{method:related_works}

\subsubsection{Unsupervised Representation Learning and Visual Clustering}
In the field of unsupervised learning, there are two main research topics: representation learning and image clustering. The former focuses on training the feature encoder by self-supervised learning. To achieve this, existing methods either design numerous pre-designed pretext tasks~\cite{doersch2015unsupervised,pathak2016context,zhang2016colorful,misra2020self} or perform contrastive learning~\cite{chen2020simple,he2020momentum,grill2020bootstrap}. Despite these efforts, these approaches are mainly used for pretraining. As an alternative, image clustering methods simultaneously optimize clustering and representation learning. Previous methods train the model using the pseudo labels derived from the most confident samples~\cite{chang2017deep,xie2016unsupervised,park2021improving,chen2020improved}, or through cluster re-assignments~\cite{caron2018deep,caron2019unsupervised}.
Recently,~\cite{asano2020self,caron2020unsupervised} are proposed to apply a balanced label assignment. 
In this work, we take advantage of the idea in~\cite{asano2020self,caron2020unsupervised} but incorporate it into the student-teacher network.

\subsubsection{Supervised and Unsupervised 3D Action Recognition} To tackle skeleton-based action recognition, many RNN-based methods~\cite{kocabas2019vibe,zhang2017view,song2018spatio} and CNN-based methods~\cite{ke2017new,li2017skeleton} are carried out.
Recently, GCN-based methods~\cite{yan2018spatial,li2019spatio,liu2020disentangling,shi2019two,zhang2020semantics} have attracted increasing attention due to their outstanding performances. We adopt the widely-used ST-GCN~\cite{yan2018spatial} as the backbone.
There are also many unsupervised methods~\cite{xu2020prototypical,nie2020unsupervised,zhao2021learning,lin2020ms2l,li20213d} proposed to learn the skeleton representations.
In particular, many methods~\cite{zheng2018unsupervised,su2020predict,kundu2019unsupervised,su2021self,yang2021skeleton} utilize the encoder-decoder structure to reconstruct skeletons from the encoded features. CrosSCLR~\cite{li20213d} proposes to apply contrastive learning for skeleton representation learning.
A recent work~\cite{yang2021unik} studies the generalizability of the models by first pretraining a model on one dataset and then finetuning it on another dataset. Our task is different from the above works, as we do not require the ground-truth labels of target data to train or finetune the model.

\subsubsection{Close-set and Open-set Transfer Learning}
In (close-set) unsupervised domain adaptation (UDA)~\cite{ben2010theory,ben2007analysis,wilson2020survey}, the source and target datasets are different in styles but have an identical label set.
Many methods aim to learn domain-invariant features by adversarial learning~\cite{ganin2016domain,cui2020gradually} or explicitly reducing the distribution discrepancy~\cite{yan2017mind}. 
For example, GVB~\cite{cui2020gradually} proposes a gradually vanishing bridge layer to facilitate the adversarial training.
Yet, these methods are not suitable for the open-set problem~\cite{mekhazni2020unsupervised}.
Novel class discovery (NCD)~\cite{han2019automatically} aims to transfer knowledge between datasets with different categories but almost the same style.
UNO~\cite{fini2021unified} trains the source set with ground-truth labels and the target set with pseudo labels generated by~\cite{caron2020unsupervised}. 
Another similar task to ours is cross-domain person re-identification (CD-ReID)~\cite{wei2018person,deng2018image}. Representative methods are based on  clustering~\cite{fan2018unsupervised,zhang2019self,zheng2021group} or domain-invariant feature learning~\cite{liu2020domain,huang2019domain,lin2018multi}. Among them, a series of methods~\cite{ge2019mutual,zhao2020unsupervised,zhai2020multiple} propose a collaborative learning scheme among multiple peer networks to alleviate the effects of label noise. However, the multiple networks are just initialized differently, making them essentially different from our method. Recently, cross-domain few-shot learning (CD-FSL)~\cite{tseng2019cross,guo2020broader} has emerged, where the source and target datasets are drawn from different domains and classes.
A few methods~\cite{phoo2020self,yao2021cross,islam2021dynamic} relax the task to allow access to unlabeled target data during training. STARTUP~\cite{phoo2020self} uses the model pretrained on the source dataset to produce soft pseudo labels for target samples, and then finetunes the model using target samples and their soft labels. 

\subsubsection{Co-training} Co-training~\cite{blum1998combining} is originally proposed for semi-supervised learning. The core idea is to train two models with different views of data, where each model is trained using the other’s most confident predictions. Later studies~\cite{wang2007analyzing,balcan2005co,qiao2018deep} extend it, \eg, to support single-view data~\cite{wang2007analyzing} or deep models~\cite{qiao2018deep}. Inspired by co-training, co-teaching~\cite{han2018co} and its variants~\cite{yu2019does,wei2020combating} are proposed to deal with noisy labels in supervised learning, where two models learn to select clean data for each other. All the above methods are proposed to solve the close-set problems in the presence of at least a few labeled samples. In this work, we attempt to address an open-set problem with new challenges.

\section{Method}

\subsection{Problem Formulation}

Formally, we denote the labeled source domain as $\mathbb{D}^{l}=\left.\left(\boldsymbol{x}_n^s, y_n^s\right)\right|_{n=1} ^{N^s}$, where $\boldsymbol{x}_n^s$ and $y_n^s$ are the $n$-th training sample and its associated action label, $N^s$ is the number of source skeleton sequences. The $N^t$ unlabeled target skeleton sequences are denoted as $\mathbb{D}^t = \{\boldsymbol{x}_n^t|_{n=1}^{N^t}\}$, which are not associated with any label. Our goal is to mine the latent classes of $\mathbb{D}^t$, which are disjoint with that of $\mathbb{D}^s$.

\begin{figure*}[t]
\centering
\includegraphics[width=\linewidth]{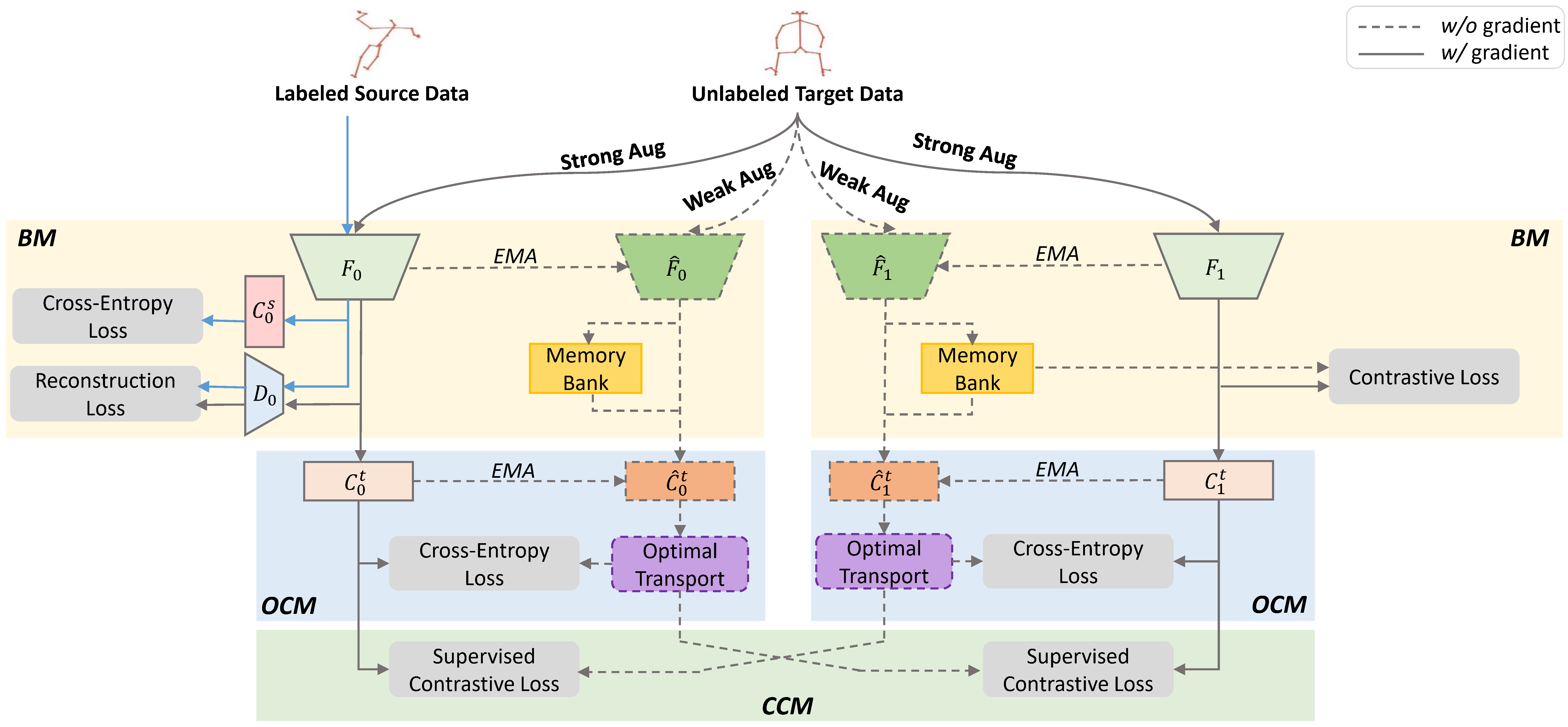}
\centering
\caption{Illustration of our CoDT network. It consists of two branches (left and right), each of which contains a base module (BM), and an online clustering module (OCM). The two branches are connected by a collaborative clustering module (CCM).}
\label{fig:arch}
\end{figure*}

\subsection{Overview}

The framework of CoDT is depicted in Fig.~\ref{fig:arch}. It is composed of two parallel branches denoted by $\mathcal{B}_0$ and $\mathcal{B}_1$, where $\mathcal{B}_0$ processes the data of both domains for exploiting domain-shared knowledge and $\mathcal{B}_1$ only processes the target data for capturing target-specific characteristics. 
Each branch contains a base module (BM) dubbed BM-$\mathcal{B}_0$ and BM-$\mathcal{B}_1$ to optimize the feature encoders. 
The BM-$\mathcal{B}_0$ is to learn discriminative information from source data via supervised learning.
To make the features more domain-invariant, we add a domain-shared decoder in BM-$\mathcal{B}_0$ for skeleton reconstruction.
The BM-$\mathcal{B}_1$ is to learn semantic similarity of target data with contrastive learning. Then we aim to collaboratively train the two branches and encourage their agreement on the clustering of target data. Yet, there are two problems: 1) How to optimize the feature clustering? 2) How to achieve co-training?  To address the problems, we propose the online clustering module (OCM) and collaborative clustering module (CCM), respectively.

\subsection{Base Module}

\subsubsection{BM-$\mathcal{B}_0$} There are two feature encoders dubbed $F_0$ and $\hat F_0$.
The $\hat F_0$ is updated by an exponential moving average (EMA) of $F_0$ and used by the subsequent OCM. 
The encoder $F_0$ first embeds the input skeleton sequences into features. The features of source samples are then fed into a source classifier $C_0^s$ and optimized by cross-entropy loss with annotated labels
\begin{equation}
\begin{aligned} 
\mathcal{L}_{sup} = \frac{1}{n^s} \sum_{i=1}^{n^s} \operatorname{CE} \left( C^s_0(F_0(\boldsymbol{x}_i^s)), y^s_i  \right),
\label{eq:target_source}        
\end{aligned}
\end{equation}
where $n^s$ is the mini-batch size of source data, and $\operatorname{CE}$ is the short of Cross Entropy loss. The decoder $D_0$  reconstructs the input skeletons from the features for both source data and target data. The reconstruction loss is computed as 
\begin{equation}
\begin{aligned} 
\mathcal{L}_{dec}  =  \frac{1}{n^s} \sum_{i=1}^{n^s} \operatorname{MSE} \left( D_0(F_0(\boldsymbol{x}_i^s)), \boldsymbol{x}^s_i  \right) + \frac{1}{n^t} \sum_{i=1}^{n^t} \operatorname{MSE} \left( D_0(F_0(\boldsymbol{x}_i^t)), \boldsymbol{x}_i^t  \right),
\label{eq:target_mse}        
\end{aligned}
\end{equation}
where $n^t$ denotes the mini-batch size of target data, and $\operatorname{MSE}$ denotes Mean Square Error loss. 
The reconstruction enforces the representations to retain generic and meaningful human structure, which can strengthen the domain invariance of representations~\cite{ghifary2015domain}. Note that, based on the principle of 'domain-sharing', when the source data and target data have different joints, we only keep their shared joints in $\mathcal{B}_0$.

\subsubsection{BM-$\mathcal{B}_1$}
To learn the target-specific representations from the target set, we employ the popular contrastive learning based on instance discrimination~\cite{he2020momentum}, where the features of the different augments of a sample are pulled together and the features of different samples are pushed apart.

Specifically, similar to BM-$\mathcal{B}_0$,
BM-$\mathcal{B}_1$ contains a feature encoder $F_1$ and an EMA encoder $\hat F_1$. 
For a target sample, we first transform it to two augments $\left(\boldsymbol{x}^t_i, \hat{\boldsymbol{x}}^t_i \right)$ by data augmentation, and then pass them into $F_1$ and $\hat{F_1}$, respectively. Here we denote the outputs as $\boldsymbol{z}_i^t= F_1(\boldsymbol{x}^t_i) $, $\hat{\boldsymbol{z}}_i^t =\hat F_1(\hat{\boldsymbol{x}}^t_i) $. Besides,
to enlarge the number of negative samples, following~\cite{he2020momentum,li2021cross},
we maintain a memory bank
$\boldsymbol M^t = \{\hat{\boldsymbol{z}}_m^t \}|_m^{N^t}$ to store the features from $\hat F_1$ of all target samples. In each iteration, the $\boldsymbol M^t$ is updated by the $\hat{\boldsymbol{z}}^t_i$ in current mini-batch, and then we compute the contrastive loss as
\begin{equation}
\begin{aligned} 
\mathcal{L}_{cont} =  \frac{1}{n^t} \sum_{i=1}^{n^t} -\log \frac{e^{\rho \cdot \mathrm{cos}(\boldsymbol{z}_i^t,\hat{\boldsymbol{z}}_i^t)}}{\sum_{m=1}^{N^t}{e^{\rho \cdot \mathrm{cos}(\boldsymbol{z}_i^t,\hat{\boldsymbol{z}}_m^t)}}  } ,
\label{eq:target_con}        
\end{aligned}
\end{equation}
where $\mathrm{cos}(\cdot)$ denotes the cosine similarity function and $\rho$ is the temperature.

\subsection{Online Clustering Module} \label{method:OCM}

\subsubsection{Pipeline of OCM}
In both $\mathcal{B}_0$ and  $\mathcal{B}_1$, we employ an OCM to guide the feature clustering via pseudo labels.
The OCM is based on the student-teacher framework~\cite{laine2016temporal,tarvainen2017mean,xie2020self}, where a teacher model (\eg, $\hat{F_0}$, $\hat{F_1}$) is updated by EMA of a student model (\eg, $F_0$, $F_1$). 
To learn the optimal cluster assignment, we add a target classifier to both student and teacher. The classifiers are denoted by $C^t_0$ and $\hat{C}^t_0$ in $\mathcal{B}_0$, and $C^t_1$ and $\hat{C}^t_1$ in $\mathcal{B}_1$. 
Following image clustering~\cite{xie2016unsupervised,ji2019invariant,van2020scan}, we set the numbers of categories of the classifiers to the number of ground-truth categories for the purpose of evaluation\footnote{In fact, as proven in~\cite{van2020scan}, even if the exact number of ground-truth categories is unknown, we can overcluster to a larger amount of clusters.}. Given a target sample $ \boldsymbol x^t_i$, inspired by~\cite{sohn2020fixmatch}, a strong augmentation operator $\mathop{A}$ and a weak augmentation operator $\mathop{a}$ are applied to transform $\boldsymbol x^t_i$ into $\mathop{A}(\boldsymbol x^t_i)$ and $\mathop{a}(\boldsymbol x^t_i)$, respectively. The teacher generates the pseudo label based on its prediction on the $\mathop{a}(\boldsymbol x^t_i)$.
Then the student is trained to predict the pseudo label by feeding $\mathop{A}(\boldsymbol x^t_i)$. These practices can enhance the invariance of representations to varying degrees of transformations. 

\subsubsection{Pseudo Label Generation}
Before elaborating on the details, we first review the objective of a classification task. Specifically, given a dataset with $N$ instances $\{\boldsymbol{x}_n\}|^N_{n=1}$
drawn from $K$ classes, the task is to maximize the mutual information between labels and input data through minimizing the Kullback–Leibler divergence $D( \boldsymbol Q ||  \boldsymbol P)$ between the model's predictions $ \boldsymbol P$ and labels $ \boldsymbol Q$: 

\begin{equation}
\begin{aligned} 
D( \boldsymbol Q ||  \boldsymbol P) =\frac{1}{N} \sum_{n=1}^{N} \sum_{y=1}^{K}   Q_{yn}\log \frac{  Q_{yn}}{  P_{yn}} 
 =\frac{1}{N} \sum_{n=1}^{N} \sum_{y=1}^{K} \big(\underbrace{ Q_{yn} \log  Q_{yn} }_{-H(\boldsymbol Q)} \underbrace{-  Q_{yn} \log P_{yn}}_{E(\boldsymbol Q || \boldsymbol P)}  \big),
\label{eq:kl}
\end{aligned}
\end{equation}
where $ Q_{yn}$ is the ($y, n$) element of $ \boldsymbol Q \in \mathbb{R}^{K \times N}$ denoting the label probability of $\boldsymbol{x}_n$ being assigned to the label $y \in \{1, \dots, K\}$. $ P_{yn}$ is the element of $\boldsymbol P \in \mathbb{R}^{K \times N}$ that denotes the probability predicted by the model. 
$D( \boldsymbol Q ||  \boldsymbol P)$ can be split into a cross-entropy term $E( \boldsymbol Q ||  \boldsymbol P)$ and an entropy term $H( \boldsymbol Q)$. 
Unlike the supervised task where $\boldsymbol Q$ is deterministic, labels are unavailable in our case, and we have to minimize $D( \boldsymbol Q || \boldsymbol P)$ \textit{w.r.t.} both the $ \boldsymbol Q$ and $ \boldsymbol P$. Further, we extend it to the student-teacher framework. Formally, it is to alternate the following steps
\begin{numcases}{}
\hat{\boldsymbol Q} \leftarrow \mathop{\arg\min}_{\hat{\boldsymbol Q} } D(\hat{\boldsymbol Q} || \hat{\boldsymbol P} ), \label{eq:step_1} \\
\boldsymbol{\theta} \leftarrow  \boldsymbol{\theta} - \epsilon \frac{\partial \big( D( \hat{\boldsymbol Q} || \boldsymbol P)  + \mathcal{L} \big) }{\partial  \boldsymbol{\theta}}, \label{eq:step_2} \\ 
\hat{\boldsymbol{\theta}} \leftarrow \alpha \hat{ \boldsymbol{\theta}} + (1 - \alpha)  \boldsymbol{\theta},
\label{eq:step_3}  
\end{numcases}      
where $\boldsymbol P$ and $ \boldsymbol{\theta}$ are the predictions and parameters of student. ${\hat{\boldsymbol P}}$ and $\hat{ \boldsymbol{\theta}}$ correspond to teacher. ${\hat{\boldsymbol Q}}$ represents the pseudo labels. 
First, ${\hat{\boldsymbol Q}}$ is calculated by Eq.~\eqref{eq:step_1}. Then, we fix $\hat{\boldsymbol Q}$ and optimize the predictions $\boldsymbol P$ of student by minimizing $D({\hat{\boldsymbol Q}} || \boldsymbol P)$ along with other auxiliary loss $\mathcal{L}$ (\eg, $\mathcal{L}_{sup}$, $\mathcal{L}_{cont}$) on a mini-batch via Eq.~\eqref{eq:step_2}, where $\epsilon$ is the learning rate of gradient descent. Finally, the teacher is updated by an EMA of the student in Eq.~\eqref{eq:step_3}, where $\alpha$ is the decay rate. 

Let's first take a look at Eq.~\eqref{eq:step_1}. Since we only consider the one-hot pseudo labels, \ie, ${\hat{ Q}}_{yn} \in \{0, 1\}$, then $H({\hat{\boldsymbol Q}}) \equiv 0 $ and $D({\hat{\boldsymbol Q}} || \hat{\boldsymbol P}) \equiv E({\hat{\boldsymbol Q}} || \hat{\boldsymbol P})$. Thus we can obtain the solution to Eq.~\eqref{eq:step_1} by taking the index that has the maximum value in the prediction of teacher, \ie,
\begin{equation}
\begin{aligned} 
{\hat{ Q}}_{yn}^* =\delta\big(y- \mathop{\arg\max}_k {\hat P}_{kn}\big),
\label{eq:fix}
\end{aligned}
\end{equation}
where $\delta(\cdot)$ is the Dirac delta function with $\delta(\cdot)=0$ except $\delta(0)=1$. The Eq.~\eqref{eq:fix} is similar in spirit to the semi-supervised learning method~\cite{sohn2020fixmatch}. 
Obviously, if $\mathcal{L} = 0$, there is a shortcut to minimize both $D({\hat{\boldsymbol Q}} || {\hat{\boldsymbol P}})$ and  $D( \hat{\boldsymbol Q} || \boldsymbol P)$ by assigning all samples to a single label, which is known as clustering degeneration.
This phenomenon can be avoided in~\cite{sohn2020fixmatch}, since the supervised training on a few labeled samples can regularize the model via $\mathcal{L}$. However, in our case, the $\mathcal{L}$ is either the supervised loss of source dataset with a distribution different from that of target dataset, or the contrastive loss for instance discrimination rather than clustering. In other words, there are discrepancies between the objectives of $\mathcal{L}$ and the ultimate goal (\ie, target sample clustering). It renders existing $\mathcal{L}$ unable to prevent clustering degeneration, as proven in Sec.~\ref{exp:OCM}.
To avoid clustering degeneration, we propose to constrain the distribution of clusters to be as uniform as possible, \ie, all clusters contain the same number of samples. 
Achieving it within Eq.~\eqref{eq:step_2} in an end-to-end fashion is difficult as the constraint involves the global distribution of the entire dataset.
Alternatively, we propose to balance the pseudo labels generated by Eq.~\eqref{eq:step_1} without gradient descent. 
Formally, we restrict ${\hat{\boldsymbol Q}}$ to be an element of transportation polytope~\cite{cuturi2013sinkhorn}
\begin{equation}
\begin{aligned}
\mathcal{U}  = \{{\hat{\boldsymbol Q}} \in \mathbb{R}_{+}^{K \times N} | {\hat{\boldsymbol Q}}^{\top} \mathbf{1}_K= \mathbf{1}_N, {\hat{\boldsymbol Q}} \mathbf{1}_N=\frac{N}{K}  \mathbf 1_K \}, 
\end{aligned}
\label{eq:polytope}
\end{equation}
where $\mathbf{1}_K$ denotes the vector of ones with dimension $K$. The Eq.~\eqref{eq:polytope} indicates that each class has equal number (\ie, $\frac{N}{K}$) of samples. Inspired by~\cite{asano2020self,caron2020unsupervised}, solving Eq.~\eqref{eq:step_1} subject to Eq.~\eqref{eq:polytope} can transformed to an optimal transport problem for mapping the $N$ data points to the $K$ centers, whose solution is 

\begin{equation}
{\hat{\boldsymbol Q}}^* = N \text{diag} (\mathbf{u}) {( \hat{\boldsymbol P}/{N})} ^ \xi \text{diag} (\mathbf{v}),
\label{eq:dot-product}
\end{equation}
where $\xi $ is a pre-defined scalar, and $ \mathbf{u} \in \mathbb{R}^K$, $ \mathbf{v}\in \mathbb{R}^N$ are two vectors computed by  the fast version of Sinkhorn-Knopp algorithm~\cite{cuturi2013sinkhorn}. Please refer to Supplementary for more details. 

Since it is costly to calculate $\hat{\boldsymbol P}$ from scratch each iteration, we maintain the same memory bank as that used in contrastive learning, and the extra cost of computing predictions from the bank is affordable.

\subsubsection{Objective of OCM} Thus, for a target instance $\boldsymbol{x}^t_i$, we can obtain its pseudo labels denoted by $\hat{y}^t_{0,i}, \hat{y}^t_{0,i} \in \{1, \dots, K\}$ in $\mathcal{B}_0$ and $\mathcal{B}_1$, respectively. Then we use the pseudo labels to train the corresponding student models by minimizing following loss function
\begin{equation}
\begin{aligned} 
\mathcal{L}_{ocm} =  \frac{1}{n^t} \sum_{i=1}^{n^t} \big( \operatorname{CE} (  \hat{C}^t_0(\hat{F}_0(\mathop{A}(\boldsymbol{x}^t_i)))
, \hat{y}^t_{0,i})  + \operatorname{CE}(\hat{C}^t_1(\hat{F}_1(\mathop{A}(\boldsymbol{x}^t_i))), \hat{y}^t_{1,i}) \big),
\label{eq:cluster_loss}        
\end{aligned}
\end{equation}

\subsection{Collaborative Clustering Module} \label{method:ccm}
To make use of complementarity between $\mathcal{B}_0$ and $\mathcal{B}_1$, we propose the collaborative clustering module (CCM) to allow co-training between two branches. 
Since the two branches have different parameters and regularized by different losses (\eg, $\mathcal{L}_{sup}$ for $\mathcal{B}_0$, $\mathcal{L}_{cont}$ for $\mathcal{B}_1$), the classes of the target classifiers of $\mathcal{B}_0$ and $\mathcal{B}_1$ are difficult to be matched exactly. Hence, the pseudo labels from one model cannot be used directly to train the other model. To overcome this challenge, we present a conceptually simple but practically effective co-training method. Instead of matching clusters between models, we propose to exchange the pair-wise relationships of samples across branches, which are explicitly matched between two branches and meanwhile can determine the performance of clustering.

Specifically, for two target instances $\boldsymbol x^t_i$, $\boldsymbol x^t_j$, we first define the binary pair-wise pseudo label as $ \mathcal{G}^t_{0,ij} =\delta(\hat{y}^t_{0,i}-\hat{y}^t_{0,j})$  in $\mathcal{B}_0$ and $\mathcal{G}^t_{1,ij} = \delta(\hat{y}^t_{1,i}- \hat{y}^t_{1,j})$ in $\mathcal{B}_1$, indicating that whether $\boldsymbol x^t_i$, $\boldsymbol x^t_j$ are from the same cluster (positive pair) or from different clusters (negative pair) in each branch. 
Besides, we define the pair-wise similarity of a pair of samples as the inner product of their predictions, \ie, $\mathcal{P}^t_{0,ij} = {\boldsymbol p^t_{0,i}}^\top \boldsymbol p^t_{0,j}$ and $\mathcal{P}^t_{1,ij} = {\boldsymbol p^t_{1,i}}^\top \boldsymbol p^t_{1,j}$, where $\boldsymbol p^t_{0,i}=\text{softmax}(\hat{C}^t_0(\hat{F}_0(\mathop{A}(\boldsymbol{x}^t_i)))) \in \mathbb{R}^K$ (likewise for $\boldsymbol p^t_{0,j}$, $\boldsymbol p^t_{1,i}$, $\boldsymbol p^t_{1,j}$). 
For co-training, we use $\mathcal{G}^t_{0,ij}$ and $\mathcal{G}^t_{1,ij}$ as the supervisions to optimize $\mathcal{P}^t_{1,ij}$ and $\mathcal{P}^t_{0,ij}$, respectively, aiming to make the similarities of positive/negative pairs increase/diminish.
To achieve it, we design the objective function similar in spirit to the supervised contrastive loss~\cite{khosla2020supervised}:
\begin{equation}
\begin{aligned} 
\mathcal{L}_{ccm} = -\frac{1}{n^t} \sum_{i=1}^{n^t} \big(
 \frac{ \sum_{j=1}^{n^t}{ \mathcal{G}^t_{0, ij} \log{\overline{\mathcal{P}^t}_{1, ij}} }}{ \sum_{j=1}^{n^t}\mathcal{G}^t_{0,ij}} +  \frac{ \sum_{j=1}^{n^t}{ \mathcal{G}^t_{1, ij} \log{\overline{\mathcal{P}^t}_{0, ij}} }}{ \sum_{j=1}^{n^t}\mathcal{G}^t_{1,ij}} \big),
\label{eq:cotraining_loss}        
\end{aligned}
\end{equation}
where $\overline{\mathcal{P}^t}_{0,ij}$ and $\overline{\mathcal{P}^t}_{1,ij}$ are defined as $\overline{\mathcal{P}^t}_{0,ij} = \frac{\mathcal{P}^t_{0,ij}}{\sum_{j=1}^{n^t}\mathcal{P}^t_{0,ij}}$,
$\overline{\mathcal{P}^t}_{1,ij} =  \frac{\mathcal{P}^t_{1,ij}}{\sum_{j=1}^{n^t}\mathcal{P}^t_{1,ij}}$.
Note that, to maximize the similarity (\ie, the inner product of the predictions) of a positive pair, both predictions need to be one-hot and assigned to the same cluster~\cite{van2020scan}. 
This property enforces the two branches to be consistent in the pair-wise relationships of cluster assignments, yielding consistent clustering performance on the two branches.

\subsubsection{Discussion}  Here we discuss the vital differences between our proposed co-training method and some related methods. ~\cite{caron2020unsupervised} swaps instance-wise labels of two views of the same image, forcing the (single) model to produce consistent predictions for different views. In contrast, CoDT exchanges pair-wise labels across models for co-training to utilize the complementarity of domain-shared and target-specific features. Recently,~\cite{zhao2021novel} proposes a mutual knowledge distillation algorithm across two different branches by comparing the feature similarity distribution between each instance and a queue of features. Differently, CoDT conducts co-training on the predictions of classifiers, leading to a direct optimization on the cluster assignments.

Apart from the idea of co-training, the objective function $\mathcal{L}_{ccm}$, to our knowledge, is also different from existing methods. To name a few, the differences between~\cite{khosla2020supervised} and ours include: 1)~\cite{khosla2020supervised} applies the loss on features, while our $\mathcal{L}_{ccm}$ is applied on predictions; 2) $\mathcal{L}_{ccm}$ has a simpler formula without any hyperparameter (\eg, temperature). Recently,~\cite{li2021semantic} proposes to adopt the instance-based contrastive loss on the predictions, showing promising performance in close-set domain adaption. Yet,~\cite{li2021semantic} handles the close-set tasks with known classes, and its loss is used for instance discrimination, which is evidently different from ours. $\mathcal{L}_{ccm}$ is also different from~\cite{van2020scan,han2019automatically,li2021cross} where a pair of samples is positive only when the similarity between their features meets a heuristic condition (\eg, is larger than a threshold). Note that, none of the above competitors involve co-training.

\subsection{Training and Test} \label{method:train_test}

In training phase, we first pretrain  BM-$\mathcal{B}_0$ and BM-$\mathcal{B}_1$ with the loss 
\begin{equation}
\begin{aligned} 
\mathcal{L}_{base} =  \lambda_{sup} \mathcal{L}_{sup} + \lambda_{dec}  \mathcal{L}_{dec} + \lambda_{cont}  \mathcal{L}_{cont},
\label{eq:stage1}        
\end{aligned}
\end{equation}
where $\lambda_*$ denotes the loss weight. After obtaining a good initialization, OCM and CCM are successively included for finetuning. The overall loss is
\begin{equation}
\begin{aligned} 
\mathcal{L}_{all} = \mathcal{L}_{base}  + \lambda_{ocm} \mathcal{L}_{ocm}  +  \lambda_{ccm} \mathcal{L}_{ccm}.
\label{eq:stage2}        
\end{aligned}
\end{equation}
In test phase, following~\cite{li20213d,islam2021dynamic}, we use the student models for testing. Specifically, if there is a target classifier in the model, we use its predictions as cluster assignments, otherwise, we use the spherical k-means~\cite{hornik2012spherical} to cluster the features.

\section{Experiment}
\subsection{Datasets and Metrics}
\emph{NTU-60}~\cite{shahroudy2016ntu} is a 3D action recognition dataset with $60$ classes. The skeletons are shot by Microsoft Kinect V2~\cite{fankhauser2015kinect}. \emph{NTU-120}~\cite{liu2019ntu} is an extended version of NTU-60, containing $120$ classes. Here we only take the classes of NTU-120 that are not overlapped with those of NTU-60, denoted by \emph{NTU-60+}. \emph{PKUMMD}~\cite{liu2017pku} is a dataset for  temporal action detection, where the trimmed action instances have been used for action recognition~\cite{lin2020ms2l}. The skeletons are also collected by Kinect V2. There are $51$ classes different from that of NTU-60+.
Different from the above in-the-lab datasets, \emph{Skeletics}~\cite{gupta2021quo} is a carefully curated dataset sourced from real-world videos~\cite{carreira2019short}. The $3$D poses are estimated by the pose estimation method VIBE~\cite{kocabas2019vibe}. 
Three cross-domain tasks are chosen to imitate various situations:
NTU-60 $\to$ Skeletics (xview), Skeletics $\to$ PKUMMD (xview), and NTU-60+ $\to$ PKUMMD (xsub).
Let's take NTU-60+ $\to$ PKUMMD (xsub) as an example. The NTU-60+ and PKUMMD are the source and target datasets, respectively, and 'xsub' is the evaluation protocol.
The former two tasks are more challenging than the last one because the former two tasks simulate the transfer learning between an in-the-lab dataset shot by 3D sensors and a real-world dataset estimated by pose estimation algorithm, while in the last task, the two datasets are shot by similar device, and are therefore more similar in style.
In the rest of the paper, we abbreviate the three tasks as N $\to$ S, S $\to$ P, and N+ $\to$ P. For evaluation, we use three widely-used clustering performance metrics~\cite{huang2020deep}, including Accuracy (ACC),  Normalised Mutual Information (NMI), Adjusted Rand Index (ARI). Due to the space constraint, we refer readers to Supplementary for more details about datasets, metrics, implementation details.

\subsection{Comparison with Different Baselines}
\label{exp:comparison}
To verify the effectiveness of CoDT, we first compare CoDT with related baselines in Table~\ref{tab:related_methods}. 
Here the results of both branches are reported.
For fairness, we use the pretrained weights of our base modules to initialize the weights of other baselines. The 'BM-$\mathcal{B}_0$ \textit{w/o} $D_0$' means that the decoder $D_0$ is not used. Compared to  'BM-$\mathcal{B}_0$', we can see that the decoder improves the performances in most cases. We further evaluate the performance when we combine 'BM-$\mathcal{B}_0$' and 'BM-$\mathcal{B}_1$' via multi-task learning, denoted by 'BM-$\mathcal{B}_0$ + BM-$\mathcal{B}_1$'. We can see that it outperforms both 'BM-$\mathcal{B}_0$' and 'BM-$\mathcal{B}_1$', demonstrating the complementarity of 'BM-$\mathcal{B}_0$' and 'BM-$\mathcal{B}_1$'.
Thereafter, we combine BM-$\mathcal{B}_0$ with other advanced methods of unsupervised skeleton/image learning~\cite{li20213d,asano2020self}. We can see that the performances are improved. However, their performance is still far behind that of CoDT, demonstrating that multi-task framework is suboptimal for combining 'BM-$\mathcal{B}_0$' and 'BM-$\mathcal{B}_1$'.    
What's more, we re-implement some representative methods of other cross-domain tasks, including UDA~\cite{cui2020gradually}, NCD~\cite{fini2021unified}, CD-ReID ~\cite{ge2019mutual} and CD-FSL~\cite{phoo2020self}.
These methods commonly share the same feature extractor among different domains.
It can be seen that our CoDT significantly outperforms all of them.
CoDT achieves salient performances since it can 1) \textit{disentangle} the learning of domain-shared and target-specific features via a two-branch framework to fully exploit their respective characteristics; 2) \textit{coordinate} the clustering of domain-shared and target-specific features to fully utilize their complementarity. 
Especially, these two properties are more advantageous when the domain gaps are larger, since the domain-shared and target-specific features are more different and complementary. It makes our CoDT extremely superior in the former two tasks with large domain gaps.

\begin{table*}[t]
\caption{Comparison between different methods. '$\dagger$' indicates that the spherical k-means is used for clustering in these methods.}
\centering
\begin{center}
\begin{tabular}{l| ccc | ccc | ccc }
\hline
\multicolumn{1}{c|}{\multirow{2}{*}{Methods}} &
\multicolumn{3}{c|}{N $\to$ S} & 
\multicolumn{3}{c|}{S $\to$ P} & 
\multicolumn{3}{c}{N+ $\to$ P} \\
\cline{2-10}
& ACC & NMI & ARI &  ACC & NMI & ARI &  ACC & NMI & ARI \\ 
\hline 
BM-$\mathcal{B}_0$$\dagger$   & $17.9$ & $22.5$ & $6.1$ & $40.0$ & $58.9$ & $30.2$ & $54.8$ & $73.7$ & $44.3$ \\
BM-$\mathcal{B}_0$  \textit{w/o} $D_0$$\dagger$  & $16.6$ & $20.5$ & $6.1$ & $38.5$ & $55.6$ & $27.6$ &  $53.3$ & $73.4$ & $43.4$  \\ 
BM-$\mathcal{B}_1$$\dagger$ & $18.7$ & $23.3$ & $6.4$ &  $43.3$ & $58.4$ & $32.1$ &  $42.7$ & $57.4$ & $32.3$\\ 
BM-$\mathcal{B}_0$ + BM-$\mathcal{B}_1$$\dagger$  & $19.4$ & $22.8$ & $6.4$ & $47.7$ & $62.1$ & $36.9$ &  $58.3$ & $74.0$ & $47.7$ \\ 
BM-$\mathcal{B}_0$ + CrossCLR~\cite{li20213d}$\dagger$  & $20.5$ & $24.1$ & $7.0$ & $49.1$ & $63.8$ & $36.6$ &  $60.4$ & $75.2$  & $47.9$ \\ 
BM-$\mathcal{B}_0$ + Asano. \textit{et.al.}~\cite{asano2020self} & $21.2$ & $26.1$ & $8.4$ &  $51.4$ & $65.4$ & $38.4$ & $62.3$ & $74.4$ & $49.4$ \\
GVB~\cite{cui2020gradually}$\dagger$ & $19.3$ & $22.2$ & $6.0$ & $37.8$ & $56.9$ & $28.4$ &  $59.5$ & $75.6$ & $50.0$ \\ 
STARTUP~\cite{phoo2020self}$\dagger$ & $19.0$ & $22.1$ & $5.9$ & $48.5$ & $63.1$ & $38.1$ &  $59.1$ & $72.5$ & $47.2$ \\
MMT~\cite{ge2019mutual} & $20.8$ & $25.5$ & $7.6$ & $52.4$ & $67.3$ & $41.2$ &  $65.4$ & $76.2$ & $55.6$ \\
UNO~\cite{fini2021unified}  & $22.5$ & $26.4$ & $9.1$ & $54.1$ & $70.0$ & $43.2$ & $66.8$ & $76.9$ & $56.7$ \\
\hline 
CoDT-$\mathcal{B}_0$ & $25.0$ &  $28.0$  & $10.7$&  $\bf 59.5$ & $\bf 74.1$ & $50.0$ &  $\bf 68.2$ & $\bf 78.5$ & $\bf 58.8$ \\
CoDT-$\mathcal{B}_1$ &$\bf 25.4$ & $\bf 28.7$ & $\bf 11.3$  & $59.4$ & $73.9$ & $\bf 50.2$ & $67.8$ & $78.1$ & $58.1$ \\
\hline 
\end{tabular}
\label{tab:related_methods}
\end{center}
\end{table*}

\subsection{Ablation Study}

\subsubsection{Effectiveness of Proposed Components}
\begin{figure}[t]
\centering
        \includegraphics[width=0.6\linewidth]{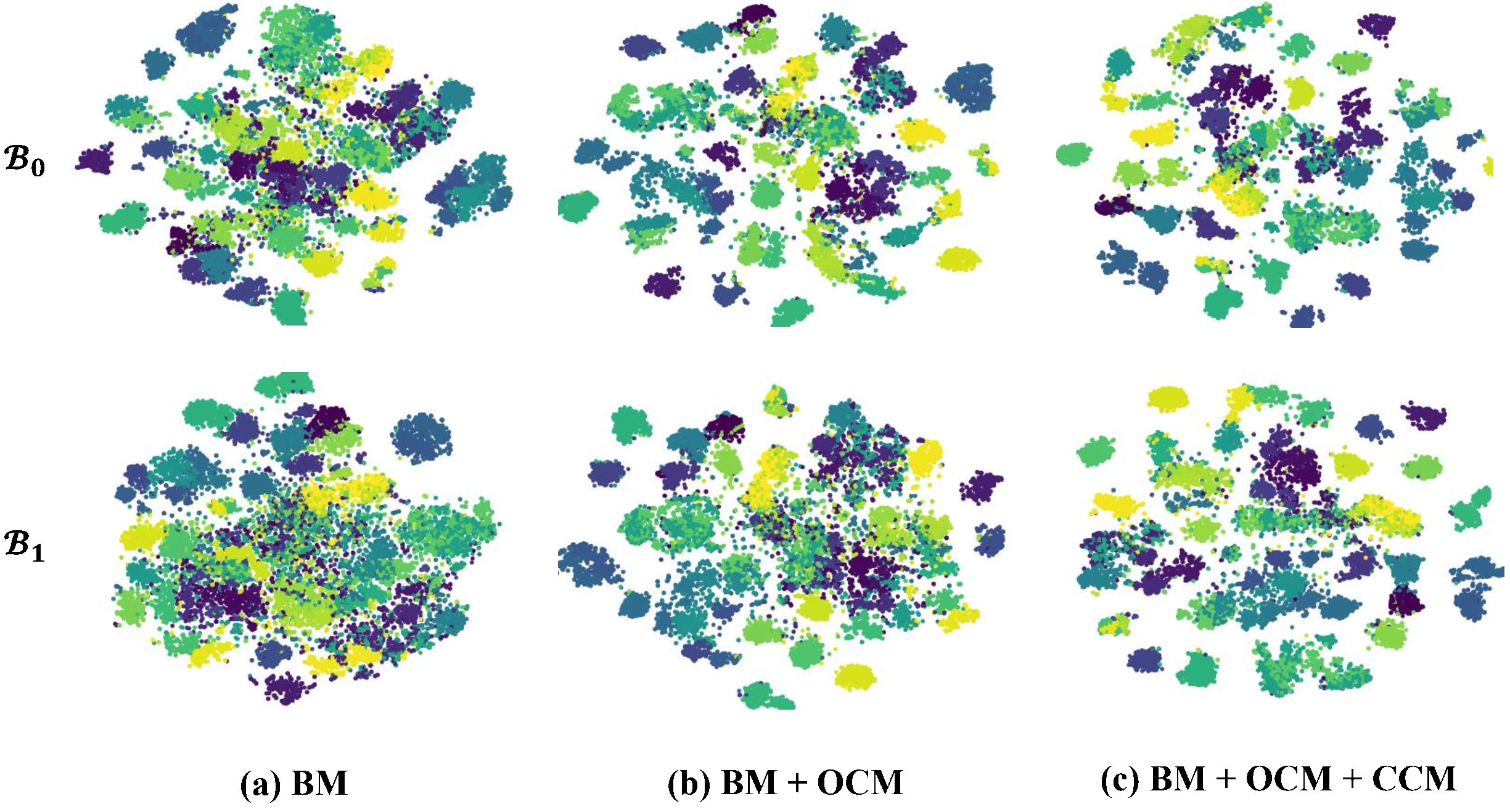}
\centering
\caption{t-SNE visualization of features learned by different methods. Different colors represent different clusters.}
\label{fig:vis}
\end{figure}

We conduct ablation studies in Table~\ref{tab:main_comp} to investigate the effectiveness of the proposed components, \ie, OCM and CCM. After introducing OCM, the performances on all branches and tasks are substantially improved when compared to the base modules. This is because the base modules are only trained to learn good representations and an offline clustering criterion (\eg, k-means) is still needed for clustering, whereas OCM is capable of optimizing both feature learning and clustering simultaneously.   
By taking account of CCM, $\mathcal{B}_0$ and $\mathcal{B}_1$ perform very closely. It is because CCM can enforce the consistency between $\mathcal{B}_0$ and $\mathcal{B}_1$ in terms of pair-wise relationships on cluster assignments. More importantly, after using CCM, the performances are greatly improved, verifying that CCM can effectively leverage the complementarity of domain-shared features and target-specific features.
In Fig.~\ref{fig:vis}, we show the evolution of the learned representations of target samples on 'N+$\to$P'. It is shown that while the clusters overlap in the beginning, they become more and more separated as the OCM and CCM are successively involved.
\begin{table}[t] 
\caption{Effect of different components.}
\centering
\begin{tabular}{l|c| ccc | ccc | ccc }
\hline
\multicolumn{1}{c|}{\multirow{2}{*}{Methods}} &
\multicolumn{1}{c|}{\multirow{2}{*}{$\mathcal{B}_*$}} & 
\multicolumn{3}{c|}{N$\to$S} & 
\multicolumn{3}{c|}{S  $\to$ P} & 
\multicolumn{3}{c}{N+ $\to$ P} \\
\cline{3-11}
&  & ACC & NMI & ARI &  ACC & NMI & ARI &  ACC & NMI & ARI \\ 
\hline 
\multirow{2}{*}{BM}$\dagger$ & $\mathcal{B}_0$ & $17.9$ & $22.5$ & $6.1$ & $40.0$ & $58.9$ & $30.2$ &  $54.8$ & $73.7$ & $44.3$ \\ 
& $\mathcal{B}_1$  & $18.7$ & $23.3$ & $6.4$ &  $43.3$ & $58.4$ & $32.1$ &  $42.7$ & $57.4$ & $32.3$ \\
\hline 
\multirow{2}{*}{BM\text{+}OCM}& $\mathcal{B}_0$  & $22.0$ & $26.8$ & $9.1$ & $55.3$ & $70.3$ & $45.1$ &  $63.7$ & $80.1$ & $53.9$ \\ 
& $\mathcal{B}_1$ & $21.8$ & $26.4$ & $8.7$ &  $54.7$ & $67.5$ & $44.3$ &  $52.6$ & $66.1$ & $41.9$ \\
\hline 
\multirow{2}{*}{BM\text{+}OCM\text{+}CCM}& $\mathcal{B}_0$ & $25.0$ & $28.0$ & $10.7$&  $59.5$ & $74.1$ & $50.0$ & $68.1$ & $78.5$ & $58.8$ \\ 
& $\mathcal{B}_1$ &$25.4$ & $28.7$ & $11.3$ & $59.4$ & $73.9$ & $50.2$ & $67.8$ & $78.1$ & $58.1$ \\
\hline 
\end{tabular}
\label{tab:main_comp}
\end{table}

\subsubsection{Analysis of OCM} \label{exp:OCM}
In Table~\ref{tab:OCM}, we give the detailed analysis of OCM, where the models with and without CCM are evaluated. Without loss of generalization, we take the task of 'S $\to$ P' as an example.  
Apart from 'ACC', we also report the metric of 'Uniformity' which is measured by the entropy of clustering results. The more uniform the distribution of clusters, the greater its value. It takes the maximum value of $3.93$ when all clusters are equal in size. 
The value corresponding to the ground-truth labels is $3.88$, indicating that the ground-truth distribution is not exactly uniform. The $\mathcal{L}$ in Table~\ref{tab:OCM} denotes the loss regularization in Eq.~\eqref{eq:step_2}, including $\mathcal{L}_{sup}$, $\mathcal{L}_{dec}$, and $\mathcal{L}_{con}$. The $\mathcal{U}$ represents the uniform constraint of Eq.~\eqref{eq:polytope}. Note that, the pseudo labels are generated by Eq.~\eqref{eq:dot-product} when $\mathcal{U}$ is used, and by Eq.~\eqref{eq:fix} otherwise.
From Table~\ref{tab:OCM}, we have following observations: 1) When neither $\mathcal{L}$ nor $\mathcal{U}$ is used, the 'Uniformity' is very small, indicating that the models encounter catastrophic clustering degeneration. 2) When $\mathcal{L}$ is used, the degeneration is alleviated and the performances increase, demonstrating that $\mathcal{L}$ is helpful to generate meaningful groups for the target data.
Besides, we observe that in the row of 'BM+OCM', the 'Uniformity' of $\mathcal{B}_1$ is larger than that of $\mathcal{B}_0$. The reason may be that the features learned by the contrastive learning have the property of uniformity in instance-level~\cite{wang2020understanding}. Nevertheless, the clusters are still not balanced enough, indicating that using $\mathcal{L}$ alone is not enough.
3) When $\mathcal{U}$ is further imposed, the clusters become more balanced, resulting in better performances. These results well prove the effectiveness of our design.
\begin{table}[t] 
\begin{minipage} {0.58\textwidth}{
\caption{Analysis of OCM, where the numbers before and after the '/' denote the results of $\mathcal{B}_0$ and $\mathcal{B}_1$, respectively. }
\label{tab:OCM}
\centering
\setlength{\tabcolsep}{0.9mm}{
\begin{tabular}{l|c c|c|c}
\hline 
Methods & $\mathcal{L}$ & $\mathcal{U}$& ACC(\%)  & Uniformity\\
\hline 
\multirow{3}{*}{BM\text{+}OCM}
& \xmark  &\xmark
  & $9.0/6.9$ & $0.70/1.09$  \\
\cline{2-5}
&  \cmark  &\xmark
 & $31.2/48.4$  & $2.76/3.68$  \\
\cline{2-5}
& \cmark  &\cmark
& $55.7/54.7$   & $3.91/3.93$  \\ 
\hline 
\multirow{3}{*}{BM\text{+}OCM\text{+}CCM}
&  \xmark  &\xmark
& $9.1/8.9$   & $0.80/0.80$    \\ 
\cline{2-5}
&\cmark  &\xmark 
& $48.4/48.7$  & $3.48/3.47$    \\ 
\cline{2-5}
&\cmark  &\cmark  & $59.5/59.4$ &  $3.91/3.93$  \\ 
\hline 
\end{tabular}
}
}
\end{minipage}
\hfil
\hfil
\hfil
\begin{minipage} {0.38\textwidth}
\caption{Comparison between our CCM with its variants, where the numbers before and after the '/' denote the ACC of $\mathcal{B}_0$ and $\mathcal{B}_1$.}

\label{tab:ccm}
\centering
\setlength{\tabcolsep}{0.4mm}{
\begin{tabular}{l| c | c}
\hline
Methods &
N $\to$ S & N+ $\to$ P  \\
\hline 
CCM-\textit{FP} & $22.3/24.2$ & $63.9/55.3$\\
\hline 
CCM-\textit{PF} & $22.8/24.7$ & $64.7/55.9$\\ 
\hline 
CCM (ours)  & $25.0/25.4$ & $68.1/67.8$\\ 
\hline 
\end{tabular}
}
\end{minipage}
\end{table}

\subsubsection{Analysis of CCM} \label{exp:CCM}
The CCM can be divided into two stages. The first is to construct pair-wise pseudo labels by comparing the pseudo labels of sample pairs in each branch, and the second is to optimize the pair-wise similarities of predictions using the pair-wise labels of the other branch. To verify the necessity of our design at each stage, we device two variants of CCM dubbed CCM-\textit{PF} and CCM-\textit{FP}, which are extended from the related methods discussed in Sec.~\ref{method:ccm}. The CCM-\textit{PF} modifies the first stage of CCM, where the pair-wise pseudo labels are constructed based on \textit{the pair-wise similarities of features} (rather than the pair-wise comparisons of pseudo labels), following~\cite{han2019automatically}. 
The CCM-\textit{PF} changes the second stage of CCM, where the pair-wise similarities of \textit{features} (rather than predictions) are optimized, following~\cite{khosla2020supervised}.
Please refer to the Supplementary for details about the variants. 
Their performances are shown in Table~\ref{tab:ccm}.
We can see that neither CCM-\textit{PF} nor CCM-\textit{FP} can ensure performance consistency between two branches, showing that in these two variants,  the cluster assignments are not fully inherited across branches. 
Due to such reason, their performances are also inferior to ours, demonstrating the effectiveness of our design.

\section{Conclusion and Future Work} \label{Conclusion}

In this paper, we propose a novel method dubbed CoDT for CD-SAR. The main idea of CoDT is to leverage the complementarity of domain-shared and target-specific features. To this end, we introduce the OCM to obtain robust pseudo labels to guide feature clustering, and the CCM to collaborate the two kinds of features. The experimental results show that our method significantly outperforms other cross-domain training or unsupervised learning methods.
In principle, our method can be adopted to other cross-domain tasks in an open-set setting (\eg, CD-FSL~\cite{phoo2020self}, CD-ReID~\cite{wei2018person}). We leave this as our future work.

~\\
\textbf{Acknowledge}
This work is supported by the National Natural Science Foundation of China under Grant No.62176246 and No.61836008.

%
%
\bibliographystyle{splncs04}
\bibliography{eccv2022submission}

\clearpage
\appendix

\title{Supplementary of CODT} 

\titlerunning{CoDT}
%
\author{ Qinying Liu \and
Zilei Wang\thanks{Corresponding Author}}
\authorrunning{Q. Liu, Z. Wang}
%
\institute{University of Science and Technology of China, Hefei, China \\
\email{lydyc@mail.ustc.edu.cn, zlwang@ustc.edu.cn}}
\maketitle

In this Supplementary, we provide more details about the optimal transport problem (Sec.~\ref{optimal_transport}), the comparison between datasets (Sec.~\ref{datasets}), the visualization of datasets (Sec.~\ref{visualization}), the evaluation metrics (Sec.~\ref{metrics}), the experimental setup (Sec.~\ref{implementation}), additional cross-domain tasks (Sec.~\ref{new_tasks}).

\section{Additional Details about Optimal Transport} \label{optimal_transport}
In the paper, we propose that generating balanced pseudo labels can be transformed to an optimal transport problem~\cite{cuturi2013sinkhorn,asano2020self}, which explores the minimum cost for assigning $N$ data points to $K$ clusters. Here we provide more details about the problem transformation and the solution. 

Specifically, based on the paper, we can express the original problem as
\begin{equation}
\small
\begin{aligned} 
\mathop{\arg\min}_{\hat{\boldsymbol Q} \in \mathcal{U} } D(\hat{\boldsymbol Q} || \hat{\boldsymbol P} ), \quad 
\text{subject to} \quad 
{\hat{\boldsymbol Q}}^{\top} \mathbf{1}_K= \mathbf{1}_N, {\hat{\boldsymbol Q}} \mathbf{1}_N=\frac{N}{K}  \mathbf 1_K,
\label{eq:objective}
\end{aligned}
\end{equation}
To see the problem more clearly, we first define that
\begin{equation}
\small
\begin{aligned} 
\boldsymbol S = \frac{1}{N}\hat{\boldsymbol Q}, \boldsymbol T = \frac{1}{N}\hat{\boldsymbol P},
\label{eq:transform}
\end{aligned}
\end{equation}
Then we can rewrite the constraint of Eq.~\eqref{eq:objective} as
\begin{equation}
\small
\begin{aligned} 
 {\boldsymbol S}^{\top} \mathbf{1}_K= \frac{1}{N}  \mathbf{1}_N, {\boldsymbol S} \mathbf{1}_N=\frac{1}{K}  \mathbf 1_K,
\label{eq:constraint}
\end{aligned}
\end{equation}
Besides, since we only consider one-hot pseudo labels, the objective function of Eq.~\eqref{eq:objective} is transformed to
\begin{equation}
\small
\begin{aligned} 
D(\hat{\boldsymbol Q} || \hat{\boldsymbol P} ) &  =   \frac{1}{N} \sum_{n=1}^{N} \sum_{y=1}^{K} -  \hat{Q}_{yn} \log \hat{P}_{yn} 
\\ & = \frac{1}{N} \sum_{n=1}^{N} \sum_{y=1}^{K} -  (N S_{yn}) \log (N T_{yn}) \\ & = 
\sum_{n=1}^{N} \sum_{y=1}^{K} -  S_{yn}(\log N + \log T_{yn})  \\ & =
- \log N \sum_{n=1}^{N} \sum_{y=1}^{K} S_{yn} - \sum_{n=1}^{N} \sum_{y=1}^{K} S_{yn} \log T_{yn}
\label{eq:objective1}
\end{aligned}
\end{equation}
According to Eq.~\eqref{eq:constraint}, $\sum_{y=1}^{K} S_{yn} = \frac{1}{N}$, $\forall n$,  then
\begin{equation}
\small
\begin{aligned} 
D(\hat{\boldsymbol Q} || \hat{\boldsymbol P}) = - \log N + \sum_{n=1}^{N} \sum_{y=1}^{K} S_{yn} (-\log T_{yn})
\label{eq:objective2}
\end{aligned}
\end{equation}
Obviously, minimizing  $D(\hat{\boldsymbol Q} || \hat{\boldsymbol P} ) $ is equivalent to  minimizing \\ 
$\sum_{n=1}^{N} \sum_{y=1}^{K} S_{yn} (-\log T_{yn})$. The $\sum_{n=1}^{N} \sum_{y=1}^{K} S_{yn} (-\log T_{yn})$ can be abbreviated as $\langle \boldsymbol S, -\log \boldsymbol T\rangle_F $, where $\langle \cdot, \cdot \rangle_F $ denotes the Frobenius dot-product. Thus, the problem of Eq.~\eqref{eq:objective} is transformed to 
\begin{equation}
\small
\begin{aligned} 
\mathop{\arg\min}_{\boldsymbol S} \langle \boldsymbol S, -\log \boldsymbol T \rangle_F, \quad \text{subject to} \quad 
{\boldsymbol S}^{\top} \mathbf{1}_K= \frac{1}{N}  \mathbf{1}_N, {\boldsymbol S} \mathbf{1}_N=\frac{1}{K}  \mathbf 1_K,
\label{eq:objective3}
\end{aligned}
\end{equation}
Using the notion of~\cite{cuturi2013sinkhorn}, we rewrite the problem as 
\begin{equation}
\small
\begin{aligned} 
\mathop{\arg\min}_{\boldsymbol S \in \mathcal{U}} \langle \boldsymbol S, -\log \boldsymbol T \rangle_F, 
\label{eq:objective4}
\end{aligned}
\end{equation}
where 
\begin{equation}
\small
\begin{aligned} 
\mathcal{U} := \{\boldsymbol S \in \mathbb{R}_{+}^{K \times N} | & {\boldsymbol S} \mathbf{1}_N= \boldsymbol r, {\boldsymbol S}^{\top} \mathbf{1}_K=\boldsymbol c \}, \\& \boldsymbol r = \frac{1}{K}  \mathbf{1}_K, \boldsymbol c = \frac{1}{N}  \mathbf{1}_N
\label{eq:polytope1}
\end{aligned}
\end{equation}
here $\boldsymbol r$ and $\boldsymbol c$ are the marginal projections of $\boldsymbol S$ onto its rows and columns, respectively. This is called an optimal transport problem~\cite{cuturi2013sinkhorn,asano2020self} between $\boldsymbol r$ and $\boldsymbol c$ given the cost matrix  '$-\log \boldsymbol T$'.
Traditional algorithms are difficult to solve it, since it involves the data points of the whole dataset.  Thus, we adopt the fast version of Sinkhorn-Knopp algorithm~\cite{cuturi2013sinkhorn} to address this issue. This amounts to adding a regularization term to Eq.~\eqref{eq:objective3}
\begin{equation}
\small
\begin{aligned} 
\mathop{\arg\min}_{\boldsymbol S \in \mathcal{U}} \langle \boldsymbol S, -\log \boldsymbol T \rangle_F +\frac{1}{\xi} D(\boldsymbol S ||\boldsymbol r {\boldsymbol c}^\top ) , 
\label{eq:objective6}
\end{aligned}
\end{equation}
here $\xi$ is a parameter that controls the balance between the convergence speed and problem approximation~\cite{asano2020self}. 
Then, the solution to Eq.~\eqref{eq:objective6} is
\begin{equation}
\small
{\boldsymbol S}^* = \text{diag} (\mathbf{u}) { \boldsymbol T} ^ \xi \text{diag} (\mathbf{v}),
\label{eq:solution}
\end{equation}
here $ \mathbf{u} \in \mathbb{R}^K$, $ \mathbf{v}\in \mathbb{R}^N$ are initially set as $\mathbf{c}$ and $\mathbf{r}$ respectively and then iteratively updated by
\begin{equation}\forall y: u_y \leftarrow\left[\boldsymbol T^{\xi} \mathbf{v}\right]_{y}^{-1} \quad \forall n: v_{n} \leftarrow\left[\mathbf{u}^{\top} \boldsymbol T^{\xi}\right]_{n}^{-1}.
\label{eq:sk}
\end{equation}
Since the $\boldsymbol S$ is relaxed to be continuous in Eq~\eqref{eq:polytope1}, we apply a rounding procedure on ${\boldsymbol S}^*$ to obtain the integral solution~\cite{asano2020self}. Besides, according to Eq.~\eqref{eq:transform}, we can see that the solution of the problem,\ie, Eq.~\eqref{eq:solution}, is exactly consistent with the formula expressed in the paper. 

\section{Additional Details about Datasets} 
\label{datasets}
The datasets used in the paper include: NTU-RGBD 60 (NTU-60)~\cite{shahroudy2016ntu}, NTU-RGBD 120 (NTU-120)~\cite{shahroudy2016ntu}, PKU Multi-Modality (PKUMMD)~\cite{liu2017pku}, Skeletics~\cite{gupta2021quo}. For NTU-60 and PKUMMD, two evaluation protocols are generally used in previous methods: 1) Cross-Subject (xsub): training data and validation data are collected from different subjects. 2) Cross-View (xview): training data and validation data are collected from different camera views. For NTU-120, two different protocols are recommended in previous methods: 1) Cross-Subject (xsub): training data and validation data are collected from different subjects. 2) Cross-Setup (xset): training data and validation data are collected from different setup IDs. Different protocols split the training and validation sets differently. 
As for Skeletics, there is only one protocol to split the training and validation sets.
Besides, Skeletics is much larger than other datasets and some classes are common with the classes of other datasets, thus we sample $30$ classes that don't overlap with the classes of other datasets. 
Three cross-domain tasks are evaluated in the paper:
NTU-60 $\to$ Skeletics, Skeletics $\to$ PKUMMD, and NTU-60+ $\to$ PKUMMD. In all the tasks, we train the models using the training split of source and target datasets, and evaluate it on the validation split of the target dataset. 
The comparison of these tasks is shown in Table~\ref{tab:datasets}.

\begin{table*}[h]
\caption{The datasets used in different tasks.}
\label{tab:datasets}
\tiny
\centering
\setlength{\tabcolsep}{0.6mm}{
 \begin{tabular}{c|c|c|c|c|c|c|c|c|c}
\hline
\multirow{3}{*}{} &
\multicolumn{3}{c|}{NTU-60$\to$Skeletics} &
\multicolumn{3}{c|}{Skeletics$\to$PKUMMD} &
\multicolumn{3}{c}{NTU-60+$\to$PKUMMD} \\
\cline{2-10}
& \multicolumn{2}{c|}{Train} &
\multicolumn{1}{c|}{Test} &
\multicolumn{2}{c|}{Train} &
\multicolumn{1}{c|}{Test} &
\multicolumn{2}{c|}{Train} &
\multicolumn{1}{c}{Test} \\
\cline{2-10}
& Source  & Target & Target  & Source  & Target & Target & Source  & Target & Target \\
\hline
Dataset & NTU-60 & Skeletics  & Skeletics & Skeletics & PKUMMD  & PKUMMD & NTU-60+ & PKUMMD & PKUMMD   \\ 
Protocol & xview & - & -& - & xview & xview  & xsub & xsub  & xsub \\
Split & train & train & val & train & train & val & train & train  & val   \\ 
Camera & fixed & moving & moving & moving & fixed  & fixed & fixed & fixed & fixed  \\
Scenario& indoor & wild  & wild & wild & indoor  & indoor & indoor & indoor   & indoor \\
Extractor & Kinect V2 & VIBE & VIBE & VIBE & Kinect V2  & Kinect V2 & Kinect V2 & Kinect V2  & Kinect V2\\
Noise& small & large & large & large& small & small & small & small & small \\
No. of Setups & $17$ & $-$ & $-$ & $-$ & $3$  & $3$ & $15$ & $3$ & $3$  \\
No. of Subjects &$40$& $-$ & $-$ & $-$ & $66$  & $66$ & $33$ & $57$ & $9$  \\
No. of Classes& $60$ & $30$ & $30$ & $30$ & $51$ & $51$ & $60$ & $41$ & $41$ \\
No. of Samples& $37,646$ & $24,265$ & $2,341$ & $24,265$ & $14,357$ & $7,188$ & $22,935$ & $18,841$ & $2,704$  \\
\hline
\end{tabular}
}
\end{table*}

\section{Visualization of Datasets} 
\label{visualization}
\begin{figure}[ht]
\centering
\includegraphics[width=\linewidth]{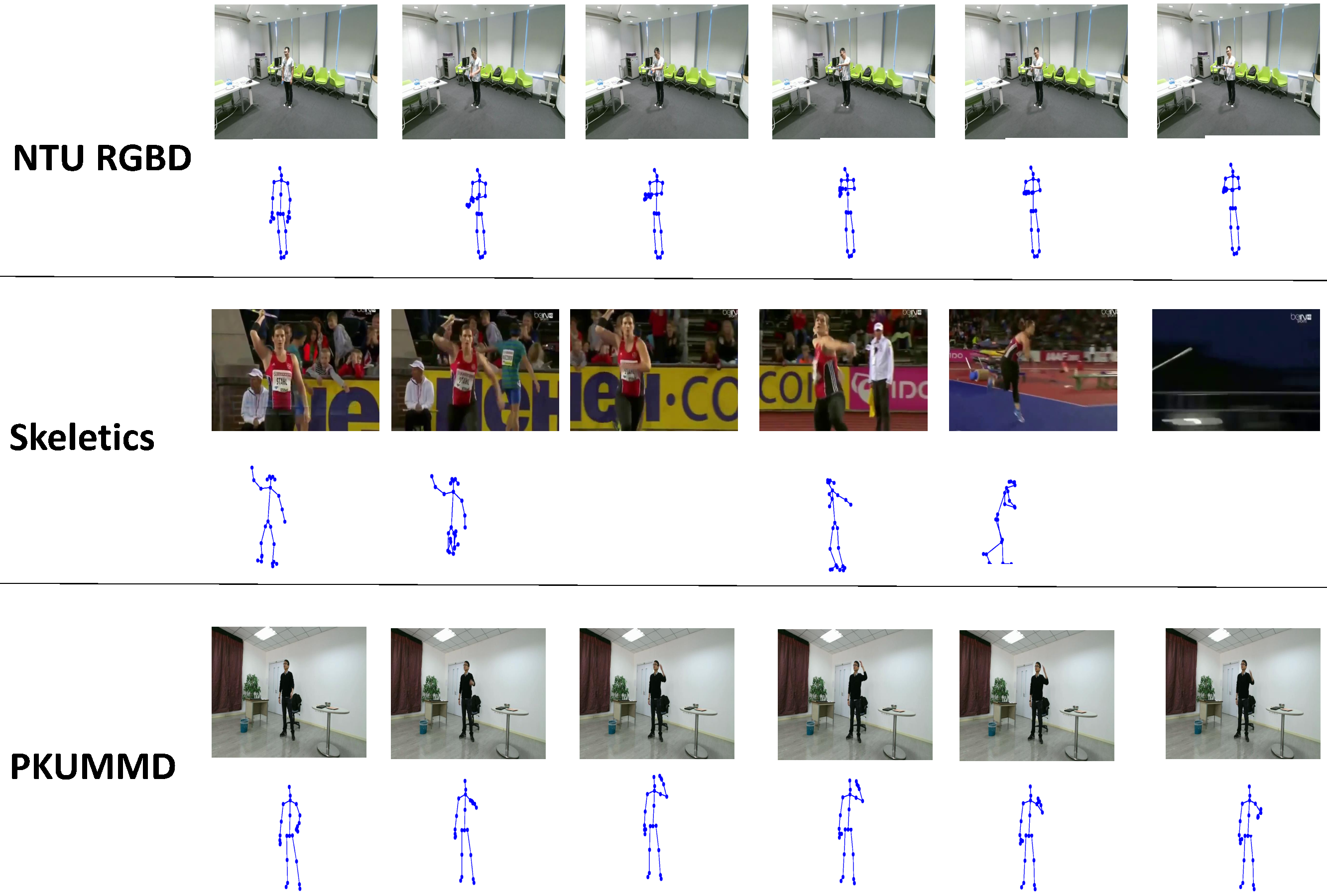}
\centering
\caption{Visualization of the image and skeleton sequences from NTU RGBD, Skeletics, PKUMMD, respectively.}
\label{fig:vis_style}
\end{figure}

\begin{figure}[ht]
\centering
\includegraphics[width=0.8\linewidth]{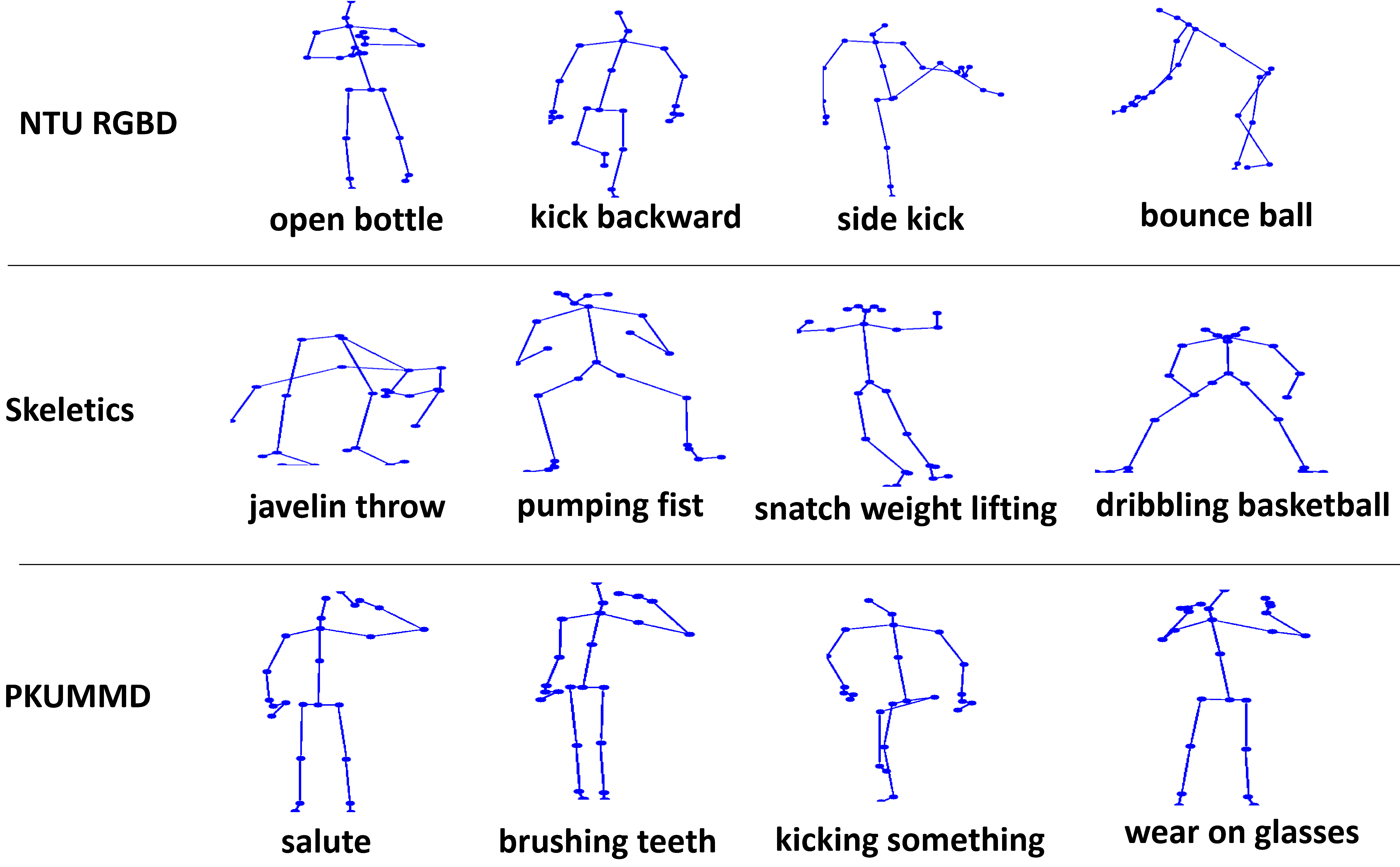}
\centering
\caption{Visualization of some specific classes in NTU RGBD, Skeletics, and PKUMMD.}
\label{fig:vis_classes}
\end{figure}

To vividly show the style differences between datasets, we provide some visualized examples in Fig.~\ref{fig:vis_style}. As can be seen from the comparison of RGB images, compared to the NTU RGBD and PKUMMD, the Skeletics is a harder dataset for skeleton extraction owing to the fast movement of humans and cameras, the incomplete body parts, the complex backgrounds, \etc.  Hence, the skeletons of NTU RGBD and PKUMMD are generally of high quality, while that of the Skeletics dataset contains more noises, \eg, missed detection, deformed body parts, as shown in Fig.~\ref{fig:vis_style}. Furthermore, we present some specific action classes for each dataset in Fig.~\ref{fig:vis_classes}.  

\begin{figure}[ht]
\centering
\includegraphics[width=0.5\linewidth]{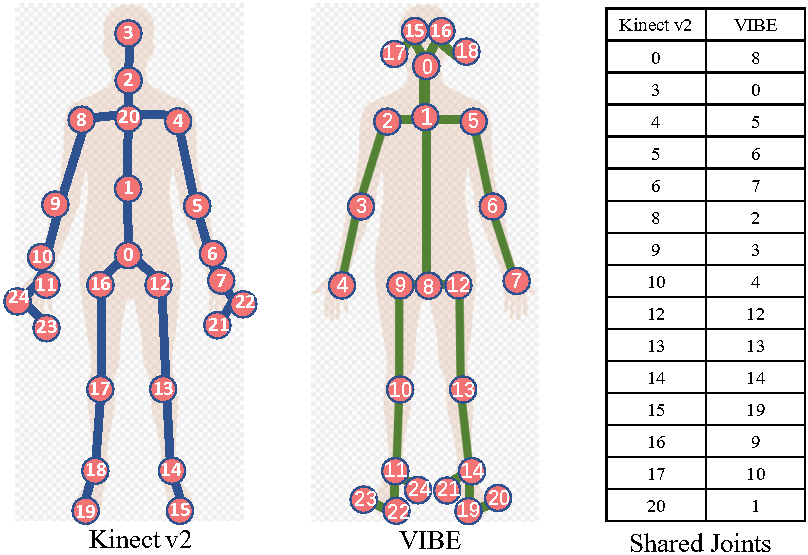}
\centering
\caption{Illustration of the $15$ shared joints of Kinect V2 and VIBE.}
\label{fig:common_joints}
\end{figure}

The joints in Microsoft Kinect V2 and VIBE are somewhat different. According to the positions of joints on the human body, we select $15$ joints as shared joints, as shown in Fig.~\ref{fig:common_joints}. Note that even these shared joints are defined slightly differently in Microsoft Kinect V2 and VIBE, which is also a kind of style gap.     

\section{Additional Details about Evaluation Metrics}
\label{metrics}
Accuracy (ACC) is computed by assigning each cluster with the dominating class label and taking the average correct classification rate as the final score. Normalised Mutual Information (NMI) quantifies the normalised mutual dependence between the predicted labels and the ground-truth. Adjusted Rand Index (ARI) evaluates the clustering result as a series of decisions and measures its quality according to how many positive/negative sample pairs are correctly assigned to the same/different clusters. All of the metrics scale from $0$ to $1$ and a higher value is better.

\section{Additional Implementation Details} 
\label{implementation}
\subsection{Implementation Details of CoDT}
The ST-GCN~\cite{yan2018spatial} is adopted as the encoder. 
All the classifiers take the encoded feature vectors as input, while the decoder $D_0$ takes the $1$D feature map before the temporal pooling layer as input. The classifiers are comprised by a full-connected layer, while the decoder is implemented by a two-layer MLP to only regress the spatial locations of joints. For data pre-processing, we resize each skeleton sequence to the length of $50$ frames by linear interpolation. Following~\cite{yao2021cross},~\textit{Shear} (\ie, spatial transformation depends on a shear amplitude $\beta$ ) and~\textit{Crop} (\ie, temporal distortion depends on a padding ratio $1 / \gamma$ ) are used for data augmentation. For weak augmentation, $\beta$ is set as $0.5$ and $\gamma$ is set as $6$. As for strong augmentation, we set $\beta$ and $\gamma$ as $1$ and $3$. Besides, we randomly mask each part of skeletons in each frame with a probability of $0.3$. The strong augmentation is only used for the student models during finetuning stage. The joint definitions are different in the Microsoft Kinect V2 and VIBE, thus in the former two tasks, we only use the shared $15$ joints of source and target skeletons in $\mathcal{B}_0$ while keeping all joints of target skeletons in $\mathcal{B}_1$. 
For the implementation of CoDT, we use SGD with momentum $0.9$ as the optimizer. In pretraining stage, the models are trained with the learning rate $0.1$. After pretraining, we first apply the spherical k-means on the features of all target samples and then use the centroids to initialize weights of target classifiers. The models are then trained with the learning rate $0.01$. The batch size $n^t$ and $n^s$ are set to $128$. The loss weights $\lambda_{sup}$, $\lambda_{dec}$, $\lambda_{cont}$, $\lambda_{cls}$ and $\lambda_{cot}$ are set to $1$, $20$, $1$, $5$ and $10$, respectively. 
The decay rate $\alpha$ of EMA is set as $0.999$. The $\xi$ is set as $10$. The temperature $\rho$ is set to $7$. All the experiments are conducted on the PyTorch~\cite{paszke2017automatic}. 

\subsection{Implementation Details of CCM-\textit{FP} and CCM-\textit{PF}}
In CCM-\textit{FP}, we measure the pairwise feature similarity by using the indices of feature elements rank ordered according to their magnitudes. If two samples share the same top-$k$ ($k=5$ is used) indices in their respective lists of rank ordered feature elements, the paired samples belong to a positive pair, otherwise, a negative pair. Note that, the above operations are performed on the features from the teacher model. 
Then the constructed binary pair-wise pseudo labels of each branch are used to optimize the pair-wise prediction similarities for the other branch, which is the same as CCM.  

In CCM-\textit{PF}, we construct the binary pair-wise pseudo labels in the same way as CCM. But different from CCM which treats the pair-wise prediction similarity as the similarities of samples,  CCM-\textit{PF} uses the cosine similarities of features as the pair-wise similarities of samples. That's, the pair-wise pseudo labels of each branch are used to optimize the pairwise feature similarities for the other branch. 
For the sake of fairness, we adopt our proposed formula of the supervised contrastive loss function to optimize both CCM-\textit{FP} and CCM-\textit{PF}.  

\section{Additional Cross-Domain Tasks} 
\label{new_tasks}
Apart from the above-mentioned cross-domain tasks, here we introduce two more tasks: FineGym (VIBE) $\to$ Skeletics (VIBE),  Kinetics-400 (OpenPose) $\to$ NTU-60 (HRNet).  
The FineGym dataset~\cite{shao2020finegym} is a fine-grained action recognition dataset with $29K$ videos of $99$ fine-grained action classes. The Kinetics-400 dataset~\cite{kay2017kinetics} contains around $300, 000$ video clips retrieved from YouTube. The videos cover as many as $400$ action classes. The VIBE~\cite{kocabas2019vibe} is a 3D pose estimator. 
The Openpose~\cite{cao2017realtime}, HRNet~\cite{sun2019deep} are two 2D pose estimators which produce 2D locations and confidence scores for the joints. That's, each joint is represented by a (pseudo) 3D tensor containing two coordinates and a score.
The poses of FineGym (VIBE) and NTU-60 (HRNet) are provided by~\cite{duan2021revisiting}, and the poses of Kinetics-400 (OpenPose) are provided by~\cite{yan2018spatial}. 
These tasks can simulate the transfer learning between datasets extracted by different 2D or 3D pose estimation algorithms. In the task of FineGym (VIBE) $\to$ Skeletics (VIBE), the FineGym dataset only contains the fine-grained gymnastic action categories, while the categories of Skeletics are much more coarse and unconstrained. And in the task of Kinetics-400 (OpenPose) $\to$ NTU-60 (HRNet), the extracted joints in two domains are extensively different in qualities and styles.  Intuitively, these two tasks are extremely difficult. 

We conduct the ablation study on these two tasks to further examine the effectiveness of our method. The results are shown in Table~\ref{tab:tasks}. It can be seen that the new tasks are very hard. For example, the ACC on the task of Kinetics-400 $\to$ NTU-60 is only around $10\%$.  
However, even in such cases, our proposed OCM and CCM can significantly improve the performances of base modules, proving the generalization ability of our method.

\begin{table*}[ht] 
\caption{Results on the additional tasks. The 'F $\to$ S' and 'K $\to$ N' represent 'FineGym (VIBE) $\to$ Skeletics (VIBE)' and  'Kinetics-400 (OpenPose) $\to$ NTU-60 (HRNet)', respectively.
}
\label{tab:tasks}
\centering
\begin{center}
\begin{tabular}{l|c| ccc | ccc }
\hline
\multicolumn{1}{c|}{\multirow{2}{*}{Methods}} &
\multicolumn{1}{c|}{\multirow{2}{*}{$\mathcal{B}_*$}} & 
\multicolumn{3}{c|}{F $\to$ S} & 
\multicolumn{3}{c}{K $\to$ N} \\
\cline{3-8}
&  & ACC & NMI & ARI &  ACC & NMI & ARI \\ 
\hline 
\multirow{2}{*}{BM}$\dagger$ & $\mathcal{B}_0$ & $16.6$ & $19.1$ & $5.0$ & $12.2$ & $26.2$ & $5.2$ \\ 
& $\mathcal{B}_1$  & $18.7$ & $23.3$ & $6.4$ & $10.4$ & $18.6$ & $3.7$  \\
\hline 
\multirow{2}{*}{BM + OCM}& $\mathcal{B}_0$  & $17.7$ & $20.1$ & $5.5$ & $13.6$ & $29.4$ & $6.2$ \\ 
& $\mathcal{B}_1$ & $21.8$ & $26.4$ & $8.7$ &  $12.1$ & $22.9$ & $5.2$ \\
\hline 
\multirow{2}{*}{BM + OCM + CCM}& $\mathcal{B}_0$ & $22.6$ & $25.4$ & $10.7$&  $14.6$ & $30.9$ & $7.2$ \\ 
& $\mathcal{B}_1$ &$23.4$ & $26.3$ & $11.0$ & $13.9$ & $29.8$ & $6.5$ \\
\hline 
\end{tabular} 
\end{center}
\end{table*}
\end{document}